\documentclass{article}
\usepackage{url}
\usepackage[accepted]{icml2021}
\usepackage[utf8]{inputenc}
\usepackage{soul}
\usepackage{multirow}
\usepackage{xstring}
\usepackage{xspace}
\usepackage{microtype}
\usepackage{graphicx}
\usepackage{subcaption}
\usepackage{booktabs} 
\usepackage{hyperref}
\usepackage{amsmath}
\usepackage{icml2021}
\usepackage{blindtext}
\usepackage{color}
\usepackage{xcolor}
\usepackage{listings}
\usepackage{csquotes}
\usepackage{tikz}
\usepackage{tabularx}
\usepackage{totpages}
\usepackage{alltt}
\usepackage{amssymb}
\usepackage{amsthm}
\usepackage{balance}
\usepackage{titling}
\usepackage{multicol}

\theoremstyle{definition}

\makeatletter
\renewcommand{\section}{\@startsection {section}{1}{\z@}{-2.3ex plus-1ex minus -.2ex}{1.6ex plus.2ex}{\reset@font\large\bf}}
\renewcommand{\subsection}{\@startsection {subsection}{2}{\z@}{-2.1ex plus-1ex minus -.2ex}{1.3ex plus.2ex}{\reset@font\large\bf}}
\renewcommand{\subsubsection}{\@startsection {subsubsection}{3}{\z@}{-2.0ex plus-0.6ex minus -.2ex}{0.8ex plus.2ex}{\reset@font\normalsize\bf}}
\makeatother

\usetikzlibrary{calc,arrows,automata,positioning}
\usetikzlibrary{decorations.pathreplacing,angles,quotes, bending}

\newcommand{\IE}{\textit{i.\@e.\@}\xspace}
\newcommand{\EG}{\textit{e.\@g.\@}\xspace}

\newcommand{\ensureprefix}[2]{\IfBeginWith{#2}{#1}{\ref{#2}}{\ref{#1#2}}}

\newcommand{\crossale}{{\sf\small Cross-ALE}}
\newcommand{\withinale}{{\sf\small Within-ALE}}

\definecolor{red}{rgb}{0.6,0,0} 
\definecolor{blue}{rgb}{0,0,0.6}
\definecolor{green}{rgb}{0,0.8,0}
\definecolor{cyan}{rgb}{0.0,0.6,0.6}
\definecolor{lightgray}{gray}{0.98}
\definecolor{lightblue}{rgb}{0.13, 0.67, 0.8}
\definecolor{lightorange}{RGB}{255,247,230}
\definecolor{codegreen}{rgb}{0,0.6,0}
\definecolor{codegray}{rgb}{0.5,0.5,0.5}
\definecolor{codepurple}{rgb}{0.58,0,0.82}
\definecolor{keywordcolor}{RGB}{94,20,64}
\definecolor{bluekeywords}{rgb}{0,0,1}
\definecolor{greencomments}{rgb}{0,0.5,0}
\definecolor{redstrings}{rgb}{0.64,0.08,0.08}
\definecolor{xmlcomments}{rgb}{0.5,0.5,0.5}
\definecolor{types}{rgb}{0.17,0.57,0.68}
\definecolor{KWColor}{rgb}{0.37,0.08,0.25}
\definecolor{KWColorTwo}{RGB}{0,137,180}
\definecolor{CommentColor}{rgb}{0.12,0.38,0.18}
\definecolor{StringColor}{rgb}{0.06,0.10,0.98}
\definecolor{darkred}{rgb}{0.65,0,0}
\definecolor{lightgrey}{rgb}{0.8,0.8,0.8}
\definecolor{marmalade}{RGB}{193,101,18}
\definecolor{peach}{RGB}{250,217,193}
\definecolor{lime}{RGB}{220,237,193}
{\end{list}}
\makeatletter
\def\BState{\State\hskip-\ALG@thistlm}
\makeatother

\colorlet{punct}{red!60!black}
\definecolor{delim}{RGB}{20,105,176}

\definecolor{pantone288}{RGB}{1,31,91}
\definecolor{pantone201}{RGB}{153,0,0}

\lstdefinelanguage{json}{
    basicstyle=\footnotesize\ttfamily,
    stepnumber=1,
    numbersep=8pt,
    string=[s]{"}{"},
    stringstyle=\color{pantone288},
    literate=
     *{0}{{{\color{pantone201}0}}}{1}
      {1}{{{\color{pantone201}1}}}{1}
      {2}{{{\color{pantone201}2}}}{1}
      {3}{{{\color{pantone201}3}}}{1}
      {4}{{{\color{pantone201}4}}}{1}
      {5}{{{\color{pantone201}5}}}{1}
      {6}{{{\color{pantone201}6}}}{1}
      {7}{{{\color{pantone201}7}}}{1}
      {8}{{{\color{pantone201}8}}}{1}
      {9}{{{\color{pantone201}9}}}{1}
      {:}{{{\color{punct}{:}}}}{1}
      {,}{{{\color{punct}{,}}}}{1}
      {\{}{{{\color{delim}{\{}}}}{1}
      {\}}{{{\color{delim}{\}}}}}{1}
      {[}{{{\color{delim}{[}}}}{1}
      {]}{{{\color{delim}{]}}}}{1},
}
\lstdefinelanguage{invar}{
    basicstyle=\color{pantone288}\footnotesize\ttfamily,
    stepnumber=1,
    numbersep=8pt,
    string=[s]{"}{"},
    rulecolor=\color{black},
    keywordstyle=\color{KWColor}\bfseries,
    stringstyle=\color{StringColor},
    moredelim=[is][\color{pantone201}]{\#}{\#},
    moredelim=[is][\color{greencomments}]{\^}{\^},
    morekeywords = { FILTER, GROUPBY, MATCH, SHUFFLE, CHOICE, LOCATIONS, MAP },
    literate=
     *{(}{{\color{black}(}}{1}
      {)}{{\color{black})}}{1}
      {|}{{\color{black}|}}{1}
      {,}{{\color{black},}}{1}
      {@}{{\color{black}@}}{1}
      {=}{{\color{black}=}}{1}
      {!}{{\color{black}!}}{1}
      {[}{{\color{black}[}}{1}
      {]}{{\color{black}]}}{1}
      {\~}{{\color{KWColor}\textbf{$\sim$}}}{1},
}

\lstdefinestyle{InvariantLanguage}{
  xleftmargin=0pt,
  basicstyle=\ttfamily\small,
  commentstyle=\color{CommentColor}\ttfamily\small,
  stringstyle=\color{darkred},
  keywordstyle=\color{KWColor}\bfseries,
  escapeinside={/*@}{@*/}
}

\icmltitlerunning{An Interpret-able feedback solution for AutoML systems}

\begin{document}

\twocolumn[

\icmltitle{Interpret-able feedback for AutoML systems}

\begin{icmlauthorlist}
\icmlauthor{Behnaz Arzani}{ms}
\icmlauthor{Kevin Hsieh}{ms}
\icmlauthor{Haoxian Chen}{upenn}
\end{icmlauthorlist}

\icmlaffiliation{ms}{Microsoft Research}
\icmlaffiliation{upenn}{University of Pennsylvania}

\icmlcorrespondingauthor{Behnaz Arzani}{bearzani@microsoft.com}
\icmlcorrespondingauthor{Kevin Hsieh}{kevin.hsieh@microsoft.com}
\icmlcorrespondingauthor{Haoxian Chen}{hxchen@seas.upenn.edu}

\vskip 0.3in
]




\begin{abstract}
Automated machine learning (AutoML) systems aim to enable training machine learning (ML) models for non-ML experts. A shortcoming of these systems is that when they fail to produce a model with high accuracy, the user has no path to improve the model other than hiring a data scientist or learning ML --- this defeats the purpose of AutoML and limits its adoption. We introduce an interpretable data feedback solution for AutoML. Our solution suggests new data points for the user to label (without requiring a pool of unlabeled data) to improve the model's accuracy. Our solution analyzes how features influence the prediction among all ML models in an AutoML ensemble, and we suggest more data samples from feature ranges that have high variance in such analysis. Our evaluation shows that our solution can improve the accuracy of AutoML by $7$-$8$\% and significantly outperforms popular active learning solutions in data efficiency, all the while providing the added benefit of being interpretable.

\end{abstract}

\vspace{-3mm}
\section{Introduction}
\label{sec:intro}
\vspace{-2mm}
The ML community has developed AutoML~\cite{AutoSklearn,tpot} to battle one of the main challenges that prevents ML adoption across a more diverse set of applications: lack of ML expertise among domain experts. However, when these systems fail to produce an accurate model, the user has no idea what to change or how to move forward. Therefore, despite the many advances in AutoML, non-ML experts find it challenging to use them in practice~\cite{proposal,scouts}.

When the AutoML system fails to produce an accurate model, the problem may be due to bad data, in-feasibility of the problem (the input is not a good predictor of the label), confounding variables, the need for better feature engineering, or any number of other problems. The (non-ML expert) user of these systems, however, does not have the required expertise to diagnose such problems nor to know how to address them. Therefore, it is crucial to accompany AutoML systems with a feedback mechanism that would inform the user about what they can do to get better results. 

A feedback mechanism for AutoML should be able to detect when the problem is infeasible, what features may need to be changed, and what additional data can help the accuracy of the model. It is important for the non-ML expert to be able to understand the justification for such feedback and to take specific action to address the issues identified. 

In this paper we present our solution for one specific type of feedback for AutoML: one that suggests new data points the user can add to the training data in order to improve the accuracy of the model produced by the AutoML.

The goal of this type of feedback is similar to the goal of active learning~\cite{activelearningsurvey}, which aims to minimize the \emph{labeling} of data samples to train an effective ML model. However, existing active learning solutions have two key shortcomings that limit their usability for AutoML. First, they do not provide any insight on why/how they suggest to label a particular set of data samples. Active learning today, is mostly used by data scientists who (a) understand concepts such as decision boundaries, uncertainty scores, and other ML jargon; and (b) often do not have the domain expertise to understand the significance of individual features (as opposed to domain experts).  The non-ML expert users of AutoML, on the other hand, have no way to understand why their training set is not good enough. This limits their ability to use their domain knowledge to improve the training set. Second, most active learning solutions are based on the characteristics of a \emph{single} and \emph{fixed} ML model, such as uncertainty~\cite{DBLP:conf/sigir/LewisG94}, variance reduction~\cite{DBLP:conf/icdm/ZhengP02}, expected model change~\cite{DBLP:conf/nips/SettlesCR07}, or a combination of pre-defined ML model features~\cite{DBLP:conf/nips/KonyushkovaSF17}. Such an assumption is ill-suited to AutoML systems (e.g.,~\cite{AutoSklearn}), which \emph{dynamically} create an ensemble of ML models based on the input. While it may be possible to apply some of the existing active learning strategies to ensembles (such as those produced by AutoML), these solutions are still limited by their inability to explain their results to non-ML experts.

We focus on providing an interpretable solution which is inspired by the query-by-committee (QBC) algorithm~\cite{qbc}. QBC utilizes model disagreements to suggest new data points for users to add to the training set. QBC creates a set of (sufficiently diverse) ML models, takes a candidate set of data points as inputs, and ranks them based on what fraction of the ML models agree on the label. The points with the most disagreement are presented to the user for labeling. We build on this idea through the following two insights to provide feedback for AutoML systems: 

\begin{itemize}
    \item Many AutoML systems produce an ensemble of ML models~\cite{AutoSklearn,tpot} and we can use these models instead of carefully crafting a set of models for feedback. Part of what enables these AutoML solutions to achieve accuracy on-par with a human expert is that they search over a diverse set of ML models, which make these ML models a reasonable committee in QBC. 
    \item To allow for interpretability, we use the variance observed in model agnostic interpretation solutions such as Accumulated Local Effects (ALE)~\cite{apley2017aleplot, apley2019visualizing} to identify new data points for the user to collect. ALE plots describe how features influence the prediction of a ML model on average. We return the average ALE plot for each feature along with the variance across the models within the AutoML ensemble (represented through error bars). The user can observe where the models within the AutoML ensemble disagree (when using each feature for predicting the label) and to interpret the feedback on a per-feature basis.
\end{itemize}

Our contributions in this work are as follows:

\begin{itemize}
    \item To the best of our knowledge, this is the first work that proposes a feedback mechanism for AutoML that suggests new data points to improve AutoML accuracy.
    
    \item Our feedback solution is an interpretable data suggestion algorithm which targets a domain-expert with little to no ML background. It is not only able to provide explanations for the data being suggested (that non-ML experts can understand) but also in some cases outperforms today's active learning solutions by 6-7\% with statistical significance. Thus, we take the first step in improving the usability of today's AutoML solutions.
\end{itemize}

\vspace{-3mm}
\section{Background and motivation}
\label{sec::motivation}
\vspace{-2mm}

We first describe why a feedback mechanism for AutoML systems is crucial for them to be usable in practice. We further outline the importance of such feedback being interpretable. Next, we provide some backgrounds on QBC and ALE plots, both of which are related to our automated and interpretable feedback solution for AutoML.

\subsection{The need for feedback}
\vspace{-2mm}

The motivation for AutoML systems such as AutoSklearn~\cite{AutoSklearn} and TPoT~\cite{tpot} is to enable non-ML experts to develop ML models for their domain specific problems. AutoML has the potential the address a key bottleneck of developing and deploying ML solutions, which is the lack of ML expertise.

However, the lack of a proper feedback mechanism for AutoML limits its usability: when it fails to produce a model with sufficient accuracy (the accuracy required by the domain expert for using the solution in deployment), the user is left with no follow up protocols --- they would either need to acquire a deeper understanding of ML or to revert to hiring a data scientists to assist them in model development. This defeats the purpose of the AutoML system.

We use a running example to illustrate both the need and the utility of an interpretable feedback solution for AutoML. The example is from a problem in computer networking: one where we decide whether to use the Scream congestion control protocol~\cite{johansson2015self} or not. Picking the right congestion control protocol is crucial for achieving good performance and there has been a lot of research for designing optimal congestion control protocols for different types of networks (e.g., ~\cite{yan2018pantheon,sivaraman2014experimental,winstein2013stochastic,dong2018pcc}). The Scream congestion control algorithm was designed for latency sensitive applications but may not be the best protocol to use in all network settings. Our problem is therefore to identify whether the application should use Scream or not to achieve the lowest end-to-end latency given the current network conditions.
In our example, the networking expert (who does not have ML background) provides AutoML with training data that identifies when Scream outperforms other congestion control protocols based on the network properties (bottleneck bandwidth, latency, loss rate, and number of concurrent flows). However, he finds that the AutoML model fails to predict the cases where Scream is not the best protocol. An ML expert would have been able to see the cause: the training data suffers from label imbalance. But the user is left puzzled with no recourse on how to proceed. 

We can solve the problem by providing {\em automated} feedback for AutoML systems. Such feedback can assist the user with suggestions on follow-up steps to improve the performance of the models to achieve acceptable accuracy with AutoML. AutoML targets users that are not versed in ML: the feedback solution has to be interpretable. In Section~\S\ref{sec::eval} we will show how interpret-ability can further assist AutoML users such as our networking expert.


We next provide necessary background on the relevant algorithms underlying our solution.

\subsection{Background on QBC}

Query-by-Committee (QBC) ~\cite{qbc,freund1997selective} is a popular active learning strategy that aims to minimize the data labeling cost for training. QBC proposes the use of a bag of (sufficiently diverse) ML models to label individual candidate points. The solution returns the data points which resulted in the most \emph{disagreements} across these ML models, and labeling these data points are more likely to improve ML model accuracy.


More formally, given a sample space $X$ of features in $\mathbb{R}^d$, it defines a concepts $c: X \rightarrow \{0,1\}$ as the mapping from the feature space to the label space and $\mathcal{C}$ as the set of concepts. It assumes the algorithm is seeing a finite set of inputs that are drawn from the feature space independently and at random based on a distribution $\mathcal{D}$. This finite set is the initial sequence of samples from the infinite set $\{x_1, x_2, \dots \}$. At each step the algorithm keeps track of the {\em version space} generated by the sequence of labeled samples used for training so far $(\hat{X},\hat{c})$. The version space is the set: $\{ c \mid c \in \mathcal{C} \quad \& \quad c(x_i) = \hat{c}(x_i) \quad \forall x_i \in \hat{X}\}$). The goal is to make progress by decreasing the size of the version space when adding each new training example. The work of~\cite{qbc} measures the information gain from that example and uses it to measure the utility of that data point. The work of~\cite{freund1997selective} shows the prediction error of QBC-like approaches decreases exponentially with the number of queries to the algorithm.


QBC has synergies with AutoML solutions such as AutoSKlearn~\cite{AutoSklearn} and TPoT~\cite{tpot}, which create an ensemble of ML models according to the training data. A key assumption of QBC is that the committee consists of a number of sufficiently diverse ML models, which are non-trivial to identify even for ML experts~\cite{activelearningsurvey}. However, this assumption can be met with AutoML systems that create an ensemble specifically when the models are individually strong and make uncorrelated errors~\cite{AutoSklearn}. Hence, we can re-purpose the models within the AutoML ensemble itself to form a QBC committee.


However, there are two main challenges when applying QBC as a feedback mechanism for AutoML: 



\noindent \textbf{(1) Requiring candidate data points.} QBC requires unlabeled candidate data points as input (sometimes called ``pool-based sampling"). The bag of models then votes on these data points to produce a set of points that are most promising (i.e., more likely to result in a higher accuracy if labeled). In the context of feedback for AutoML, if the user provides a set of candidate points they can bias the output based on their choice and limit the effectiveness of the data the solution produces (remember the users of this solution are {\em not} ML experts and thus cannot be relied upon to produce a representative candidate set). Alternatively, one could create a candidate set by simply sampling based on the distribution of the training set. However, the training set too can suffer from the same biases: in the context of our running example, the user may collect data from a production network and miss observing unique cases that only occur when the loss rate of the network is higher due to failures or (less-common occurrences of) congestion. Our solution does not rely on an input candidate set and provides samples from a subspace of the feature-space to the user.

\begin{figure}
  \centering
\includegraphics[width=0.38\textwidth]{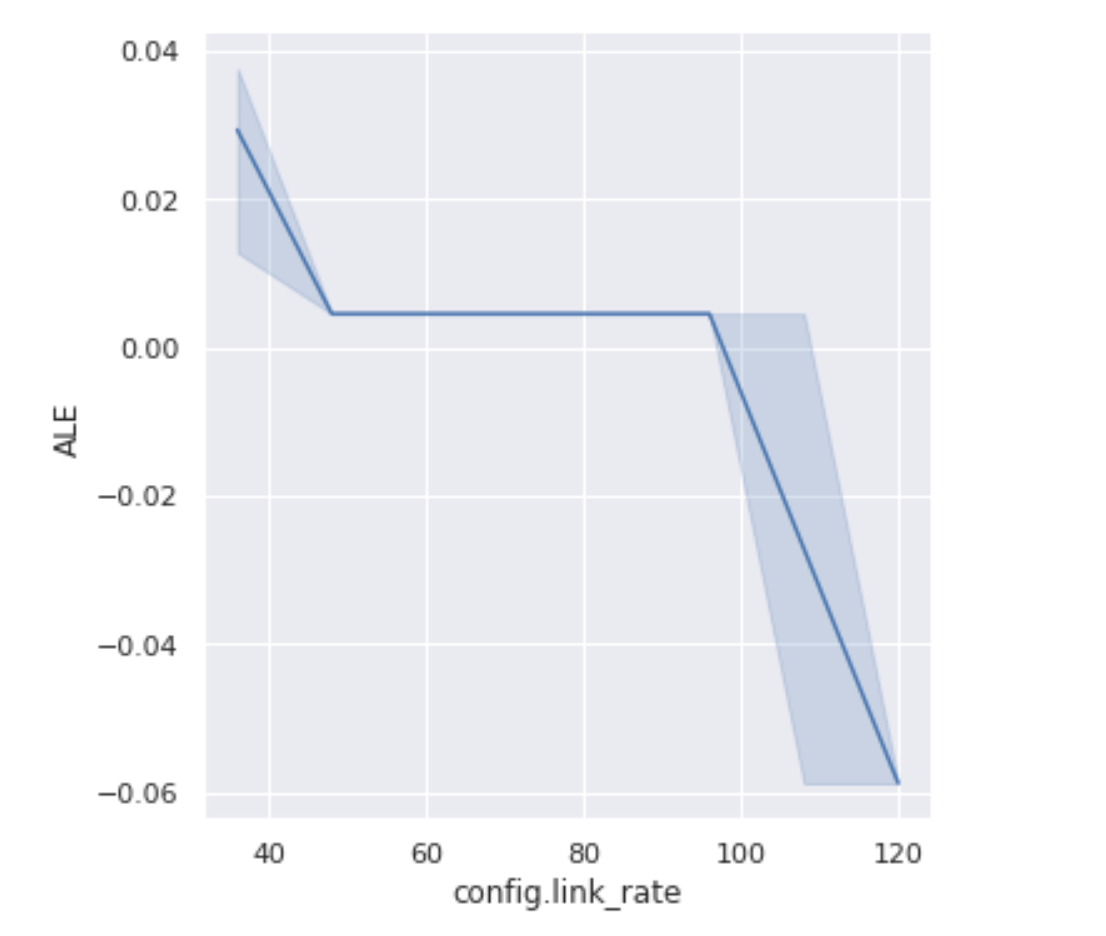}
 \caption{An ALE plot for a problem deciding whether to pick the Scream congestion control protocol vs other protocol candidates. The feature set includes: the network bottleneck bandwidth (link rate), the end-to-end latency, the network loss-rate, and the number of flows sharing the bottleneck link.}
 \label{fig:ale}
\end{figure}

\noindent \textbf{(2) They are not interpretable.} QBC and other active learning solutions are not interpretable by a non-ML expert who is unfamiliar with concepts such as decision boundaries and model confidence. Instead of using the model predictions, we use model-agnostic interpretation algorithms (such as ALE, ~\S\ref{subsec::ale}) and the variance of the interpretations as a measure of disagreement. Thus, by returning the ``aggregate interpretations'' (i.e. the average of the interpretations across the different underlying ML models) along with the standard deviation the user can understand how the features contribute to the overall disagreement among the ML models in the AutoML ensemble. Figure~\ref{fig:ale} shows ALE-based feedback (our solution) for our running example: the domain expert can see there is more disagreement in both high and low link rates across the ensemble about whether Scream is the best protocol. The feedback can help them understand the role of this feature and why they should add more high/low link rate points to the training set. 

\subsection{Background on ALE plots}
\label{subsec::ale}


Accumulate Local Effect (ALE) plots describe how features influence the prediction of a ML model on average~\cite{apley2019visualizing}. More formally, suppose we fit a supervised ML model $\hat{f}(\cdot)$ that predicts $Y$ for a $d$-dimensional feature vector $X=(X_1,X_2,...,X_d)$ based on $n$ training data $\{y_i,x_i=(x_{i,1},x_{i,2},...,x_{i,d})\}_{i=1,2,...n}$. The ALE function for a feature $X_S$ at value $x_s$ is defined as:

\vspace{-15pt}
\begin{align*}
\label{eq::ale}
  &  f_{S,ALE}(x_S) = \nonumber \\
   & \int_{x_{min,S}}^{x_S} E_{X_C\mid X_S} \big[\hat{f}^S(X_S,X_C) \mid X_S = z_S\big] \mathrm{d}Z_S - const
\end{align*}
\vspace{-15pt}

where $X_C$ represents the vector of all the features other than $X_S$ (i.e., $X_C=(X_k:k=1,2,...,d;k \ne S)$). $\hat{f}^S(x_S,x_C) \equiv \frac{\partial \hat{f}(x_S,x_C)}{\partial x_S }$ represents the local effect of $x_S$ on $\hat{f}(\cdot)$ at $(x_S, x_C)$. In practice, the calculation of $\hat{f}^S(x_S,x_C)$ is typically replaced by the differences in the predictions over an interval in a quantized grid. $x_{min,S}$ is some value chosen near the lower bound of the smallest observation of $X_S$ ($\min \{ x_{i,S} \}_{i=1,2,...,n}$). The $const$ is chosen to vertically center the plot. More details on ALE can be found in~\cite{apley2019visualizing, molnar2020interpretable}. This computation results in a figure similar to Figure~\ref{fig:ale} and allows the user to understand what the ML model has learned about that feature (while eliminating the impact of correlated features). 


\vspace{-3mm}
\section{Algorithm}
\vspace{-2mm}

The intuition for our algorithm is that the accuracy of AutoML systems such as AutoSKlearn and TPoT is in part due to their use of ensembles which contain a set of diverse ML models with uncorrelated errors~\cite{AutoSklearn}. Thus, these models are well suited for QBC-like algorithms. In order to provide interpretability, we change our metric for model disagreements from prediction disagreements to the variance across their ALE plots. ALE plots (and other model-agnostic interpretation methods) are designed to provide insights to the user about what the model has learned. Therefore, through this simple change to QBC we are able to provide additional insights to the user about why they should sample those additional data points and add them to the training set. Our solution is as follows:

    \begin{enumerate}
        \item The algorithm takes as input the ensemble produced by AutoML $\mathcal{M}$, a threshold $\mathcal{T}$ setting the variance we are willing to tolerate, the feature-set $X$ and the domain of each feature in that set: $R(X_S)$ for each $X_S \in X$ (the range of values each feature can take in $\mathbb{R}$).
        
        \item The output of this algorithm is a set $\mathcal{K}$ of data points for the user to label and add to the training set.
        
        \item For each model in $\mathcal{M}$ we apply a model-agnostic interpretation algorithm. We use ALE in this work. We can also use other algorithms such as partial dependence plots (PDP)~\cite{greenwell2017pdp} or Friedman's H-statistic~\cite{friedman2008predictive}.
        
        \item We then compute the standard deviation across the ALE values of models in $\mathcal{M}$ for each feature $X_S \in X$ in its range $R(X_S)$.  In our implementation we quantize the space $R(X_S)$ and return the variance across each quantile (this, however, is not a fundamental choice and users can opt for the continuous alternative).  
        \item Return the subspace where the standard deviation is high (higher than $\mathcal{T}$) as the region for the user to sample more points from. These subspaces are essentially a collection of hyperplanes $\cup_{i} A_i x \le b_i$ where $x$ is an $\mid X \mid \times 1$ variable vector representing the features and $A_i$ are $m \times \mid X \mid$ matrices of constants ($\mid X \mid$ is the cardinality of the set $X$).  Similarly, $b_i$ are $ m \times 1$ vectors of constants. The magnitude of $m$ depends on the regions where the standard deviation exceeds $\mathcal{T}.$ The equations in $A_i x \le b_i$ describe the regions in the ALE plot where the variance is higher than $\mathcal{T}$. Note, we need a union of $A_i x \le b_i$ subspaces because the space need not be continuous: in our example in Figure~\ref{fig:ale}, if we assume the bottleneck bandwidth is the only feature where the variance exceeds the threshold, our feedback returns $x \le 45 \cup x \ge 99$ where $x$ is the config.link\_rate feature. The user can now sample more points from these regions and add them to the training set (alternatively we can also sample the space ourselves and return a set of points to the user).
        
        
        \item Return the average ALE plots (along with error-bars) as explanations to the user (for example, see Figure~\ref{fig:ale}).
    \end{enumerate}
    
\begin{algorithm}
\SetAlgoLined
\KwResult{AutoML feedback}
 \textbf{Input:} $\mathcal{M}, \mathcal{T}, X, \mathcal{R}(X_S),n$;\\
 \textbf{Output:} $\mathcal{K} = \{K_{X_S} \mid X_S \in X \}$;\\
 \textbf{Initialize:} $K_{X_S} = \{\}$ \quad $\forall X_S \in X$;\\
 \ForEach{$X_S \in X$}
 {
 Quantize $\mathcal{R}(X_S)$ into $R_S = \{r_1,r_2,\dots r_n\}$;\\
 \ForEach{$r_i \in R_S$}{
 compute  $f_{S,ALE}(r_i) \quad \forall m \in \mathcal{M}$ ;\\
 \uIf {$std(f_{S,ALE}(r_i)) \ge \mathcal{T}$}{
 $K_{X_S} = K_{X_S} \cup \{r_i\}$;\\
 }
 }
 }
 \caption{{\withinale} algorithm. Here $\mathcal{M}$ refers to the AutoML ensemble, $\mathcal{T}$ is a pre-defined threshold, X is the set of all features, and $\mathcal{R}(X_S)$ is the range of $X_S$.}\label{algo::within_feedback}

\end{algorithm}   
 \vspace{-2mm}

\textbf{Algorithm variants.}
A variant of the algorithm above runs AutoML multiple times and uses the disagreements across the AutoML runs for feedback. This variant replaces the ensemble $\mathcal{M}$ in algorithm~\ref{algo::within_feedback} with a set $\mathcal{M}'$ where each $m \in \mathcal{M}'$ is an ensemble returned by an AutoML run on the training data. AutoML runs are intrinsically non-deterministic due to their probabilistic search algorithms and therefore, each run produces a different bag of models. Thus, this approach is more robust in that it creates a more diverse bag of models but also is more expensive as each AutoML run can take a long time. We refer to this variant as {\crossale} feedback and the earlier version as {\withinale} feedback.  The {\crossale} solution has an additional benefit: it allows us to extend our feedback solution to non-ensemble based AutoML systems. We defer an investigation into the viability of that idea to future work. In our evaluations we use $10$ AutoML runs to create $\mathcal{M}'$. 

\section{Evaluation}
\label{sec::eval}

In this section, we demonstrate how our {\em automated} and {\em interpret-able} feedback solution for AutoML can help users take action and improve the accuracy produced by the system. We first describe the experimental setup in \S\ref{sec::setup}.
Our goal is to answer the following questions:

\begin{enumerate}

\item Can our feedback solution help users improve the accuracy of AutoML systems?
\item How does our feedback solution work when the user has complete control and can collect any data the feedback solution specifies vs when the pool of data available to the feedback system is fixed?
\item How does our feedback solution compare against active learning solutions that also suggest new labeled data that can help improve the accuracy of ML models?
\item How does our solution compare against other, traditional, data-scientific approaches?
\end{enumerate}

\subsection{Experiment methodology}
\label{sec::setup}

\noindent{\textbf{Implementation.}} We use AutoSKlearn~\cite{AutoSklearn} as our AutoML platform. We implement our solution with $\sim$ 500 lines of python code. Each training experiment runs the AutoML system for an hour. We repeat each experiment 10 times, and with 20 different test sets, to ensure statistical significance across AutoML runs.

\noindent{\textbf{Datasets.}} We use two seperate datasets:

\noindent{\textit{Scream vs rest.}} This is our running example from~\S\ref{sec::motivation}. Here, a network emulator can provide target performance (label) for a given network condition (feature variables). We use the Pantheon emulator~\cite{yan2018pantheon}, which has been used for various ML tasks in computer networking~\cite{sivaraman2014experimental, yan2018pantheon}.


Because we collect the data through emulation, we can easily collect any additional data the feedback solution specifies. We use $1161$ for training the initial AutoML system and add an additional $280$ points based on the feedback. We divide an additional $4850$ data points into 20 (roughly) equally sized test sets at random (to measure statistical significance). We also collect 2000 data points uniformly at random as the candidate-set for our active learning benchmarks. 

\noindent{\textit{UCL dataset.}} We further evaluate our solution using the CoverType dataset in the UCL KDD archives~\cite{Covertype}. This is a dataset with 54 features, some of which are binary. We divide the data and use $40\%$ for training, $20\%$ for testing --- which we further divide into 20 test sets for statistical significance, and the remaining $40\%$ as our candidate feedback pool (our raw training set before feedback consists of $232404$ points). To ensure we also measure statistical significance across different data splits, we also repeat this splitting $5$ times and report the results across all these scenarios. The CoverType dataset is an example of a multi-class classification problem with $7$ possible labels.

\noindent{\textit{UCL dataset 2.}} To show the benefit of the interpretability of our solution we use the ``Internet Firewall Data Data Set'' from the UCL KDD archives~\cite{Firewall}. We use the same approach as we did for the first UCL dataset to create training, test, and feedback sets for this data set (26212 trining samples). This also is a multi-class classification problem with $4$ classes.


\noindent{\textbf{Metrics.}} We compute {\em balanced accuracy} over each test set as our accuracy measure. We use this metric (instead of accuracy) to avoid biases due to label imbalance. To measure statistical significance, we use the p-values reported by the one-sided wilcoxon signed ranked test~\cite{woolson2007wilcoxon}. We use an $\alpha = 5\%$ to reject the null hypothesis (corresponding to $95\%$ probability that we can safely reject the null hypothesis and {\em can accept the alternate hypothesis}: note that when we can't reject the null hypothesis it does not mean we can accept that the null hypothesis is true).

\begin{table*}
\begin{center}
\begin{tabular}{ l c c c c}

\textbf{Algorithm (X)} & \textbf{balanced accuracy} & \textbf{P(no feedback, X)} & \textbf{P(X, within ALE)} & \textbf{P(X, cross ALE)}\\
\hline
\hline
Without feedback & $68.7\% \pm 4.05\%$ & NA & 0.0009 & $3.33 \times 10^{-6}$\\ 
{\color{blue}{\withinale}} &  {\textbf{\color{blue}$71.2\% \pm 4.3\%$}} & {\color{blue}\textbf{$0.0009$}} & NA & {\color{blue}\textbf{$1.66 \times 10^{-5}$}} \\  
{\color{blue}{\crossale}} & {\color{blue}\textbf{$75.0\% \pm 4.4\%$}} &{\color{blue} \textbf{$3.33 \times 10^{-6}$}} & {\color{blue}0.99} &NA\\
 Uniform & $64.1\% \pm 4.1\%$ & $0.99$ & $4.02\times 10^{-5}$ & $7.86\times10^{-6}$\\
 Confidence based & $67.1\% \pm 5.5\%$ & $0.99$ & $2.38\times 10^{-7}$ & $2.38\times10^{-6}$\\
 Upsampling & $76.7\% \pm 2.7\%$ & $2.38\times10^{-7}$ & $1$ & $0.99$\\
 QBC  & $68.9\% \pm 5.1\%$ & $0.093$ & $0.004$ & $2.38\times10^{-7}$\\
 {\sf\small Within-ALE-Pool} (180 points) & $67.4\% \pm 4.9\%$ & 0.99 & $7.86 \times 10^{-6}$ & $4.76 \times 10^{-7}$\\
 {\sf\small Cross-ALE-Pool} (91 points) & $69.18\% \pm 3.9\%$ & $0.123$ & $0.013$ & $0.013$\\
 \hline
\end{tabular}
\end{center}
\vspace{-2mm}
 \caption{Scream vs rest balanced accuracy numbers. For brevity we use P(x,y) --- note the ordering of the arguments --- to indicate the p-values of the one sided Wilcoxon signed ranked test that checks the hypothesis that $x == y$ and has the alternate hypothesis $x \le y$. The {\sf\small Within-ALE-Pool} and {\sf\small Cross-ALE-Pool} algorithms are the variants of the ALE approach that use a candidate pool. All the other algorithms add 280 points to the training set.  \label{table::balanced_accuracy}} 
 \vspace{-3mm}
\end{table*}

\noindent{\textbf{Benchmarks.}} We use the following benchmarks:

\noindent{\textit{Uniform.}} We uniformly sample from the feature space the same number of points that we use for the ALE feedback and add it to the training set as the simplest baseline.

\noindent{\textit{Confidence-based feedback.}} We compare against one of the most commonly used active learning solution, confidence-based sampling (or uncertainly sampling)~\cite{DBLP:conf/sigir/LewisG94, DBLP:journals/jmlr/TongK01}. We use the prediction probability returned by AutoSKlearn as a measure of confidence and, from a uniformly sampled candidate pool (2000 uniformly sampled points for the Scream vs rest dataset), return the points with the least confidence as feedback. We fix the number of points we return to the user so that it is the same as the number of points returned by the ALE feedback.

\noindent{\textit{QBC for AutoML.}} We also compare against QBC, a classic active learning strategy. We modify QBC so that it uses the models in the AutoML ensemble as the committee instead of creating a curated ensemble which is a difficult challenge itself~\cite{DBLP:conf/nips/Gilad-BachrachNT05, activelearningsurvey}. The main diffrence between this approach and ours is in using ALE-variance instead of entropy to decide whether to add a data point to the training set. Here, we use the same candidate pool as the one we use for the confidence-based feedback.

\noindent{\textit{Upsampling.}} The Scream vs Rest dataset suffers from label imbalance. Therefore, in that instance we compare our solution to one standard data-science solution to label imbalance, upsampling~\cite{chawla2002smote}.


\vspace{-3mm}
\subsection{Results over the Scream vs. rest dataset}
\vspace{-2mm}

We first describe our results over the scream vs rest dataset (Table~\ref{table::balanced_accuracy}). To ensure the comparisons are statistically significant, we also report the statistical significance (p-values) of the one sided Wilcoxon signed ranked test where the alternate hypothesis is the hypothesis that the non-ALE approach has less balanced accuracy compared to the ALE-based approach.  We see our ALE based approaches outperform both active learning approaches (QBC and confidence based) with high statistical significance --- we can reject the null hypothesis that the distributions are equal and accept the alternate hypothesis that our feedback produces higher balanced accuracy compared to these active learning solutions.

This type of feedback is what the ALE-based feedback is designed for --- the user has the ability to gather more data based on the feedback and provide the necessary labels. Our intuition is that the ALE-based solutions outperform the other baselines because these baselines are limited to the unlabeled candidate data pool provided to them: they do not suggest new candidate points themselves but instead are only able to assign a score to any unlabeled data point in the given pool. In contrast, ALE suggests the entire subspace of samples which are likely to improve AutoML's accuracy.

To check this hypothesis we also use the same candidate pool as the one used for the active learning baselines for the {\crossale} and {\withinale} approaches (these are the {\sf\small Within-ALE-Pool} and {\sf\small Cross-ALE-Pool} algorithms in Table~\ref{table::balanced_accuracy}). Because we limit the algorithms to the candidate pool (the ALE-based approaches throw out data points that do not fall within the subspace the algorithm returns), we are not able to use the same number of points compared to the other benchmark active learning solutions, which significantly disadvantages our ALE-based approaches (we show the number of data points we add to the training set in parenthesis in Table~\ref{table::balanced_accuracy}). Indeed, the performance of the ALE-based algorithms drops significantly and becomes comparable to the other active learning solutions. We see this observation holds on the UCL dataset as well (see~\S\ref{sec::UCL}). However, even in these scenarios, the ALE approach is comparable to the other active learning solutions and is preferred as a feedback solution to AutoML due to its interpretability. 

Upsampling the training set outperforms all other approaches. This is because upsampling deals with the fundamental underlying problem with the training set (label imbalance). Even in this case the {\crossale} feedback is on average within $1\%$ of the accuracy of upsampling.

Finally, we see that, as expected, {\crossale} outperforms {\withinale}. However, using {\crossale} is more costly as the user will have to run the AutoML system multiple times. 


%


\begin{table*}
\begin{center}
\begin{tabular}{ l c c c}

\textbf{Algorithm (X)} & \textbf{balanced accuracy} & \textbf{P(no feedback, X)} &  \textbf{P(X, within ALE)}\\
\hline 
\hline
Without feedback & $88.9\% \pm 2.1\%$ & NA & $0.2$\\
{\color{blue} {\withinale}} & {\color{blue}$89.1\% \pm 1.5\%$} & {\color{blue}$0.2$} & NA\\
Uniform & $89.7\% \pm 1.0\%$ & $0.0015$ & $0.98$\\
Confidnece based & $90.5\% \pm 1.1\%$ & $8.63\times 10^{-12}$ & $0.99$\\
QBC & $89.9\% \pm 1.0\% $ & $3.01 \times 10^{-6}$ & 0.99\\
\hline
\end{tabular}
\end{center}
\vspace{-2mm}
 \caption{UCL dataset balance accuracy for comparing {\withinale} with other algorithms. We separate the {\crossale} from {\withinale} in this case to make sure the number of feedback points each algorithm uses is the same. For brevity we use P(x,y) to indicate the p-values of the one sided Wilcoxon signed ranked test that checks the hypothesis that $x == y$ and has the alternate hypothesis $x \le y$.\label{table::balanced_accuracy_UCL_within}}
 \vspace{-3mm}
\end{table*}

\begin{table*}
\begin{center}
\begin{tabular}{ l c c c}

\textbf{Algorithm (X)} & \textbf{balanced accuracy} & \textbf{P(no feedback, X)} &  \textbf{P(X, cross ALE)}\\
\hline 
\hline
Without feedback & $88.9\% \pm 2.1\%$ & NA & $9.66 \times 10^{-10}$\\
{\color{blue} {\crossale}}& {\color{blue}$90.1\% \pm 1.1\%$ }& {\color{blue}$9.66 \times 10^{-10}$} & NA  \\
Uniform & $90.8\% \pm 1.3\% $ & $1.30 \times 10^{-17}$ & $0.99$ \\
Confidence based & $91.3\% \pm 1.3\%$ & $6.17\times 10^{-20}$ & $0.99$\\
QBC & $90.22 \pm 1.0\% $ & $2.64\times10^{-11}$ & $0.67$\\
\hline
\end{tabular}
\end{center}
\vspace{-2mm}
 \caption{UCL dataset balance accuracy for comparing {\crossale} with other algorithms. We separate the {\crossale} from {\withinale} in this case to make sure the number of feedback points each algorithm uses is the same. For brevity we use P(x,y) to indicate the p-values of the one sided Wilcoxon signed ranked test that checks the hypothesis that $x == y$ and has the alternate hypothesis $x \le y$.\label{table::balanced_accuracy_UCL_cross}}
 \vspace{-3mm}
\end{table*}

\vspace{-4mm}
\subsection{Results over UCL dataset}
\label{sec::UCL}
\vspace{-2mm}

The UCL datasets disadvantage the ALE-based feedback algorithms in two ways (a) in this scenario, we are limited to a candidate pool and cannot label additional data points --- ALE does not allow us to ``rank'' individual data points (as opposed to active learning solutions that do). Hence, we can only filter the candidate pool based on the subspaces with high variance; (b) this dataset also contains features that are not continues (binary features) which also limits how many points are available in the candidate pool that also fall in the subspaces ALE-based feedback returns: the probability that a point will not fall in the ALE subspace is higher.

Tables~\ref{table::balanced_accuracy_UCL_within} and~\ref{table::balanced_accuracy_UCL_cross} show the results: despite these disadvantages the balanced accuracy achieved using the ALE-based feedback is comparable to the other active-learning solutions.


While ALE-based feedback yields similar accuracy to other active learning strategies, ALE-based feedback is still a better solution as they provide interpretations for the feedback.

\vspace{-3mm}
\subsection{Results over UCL dataset 2}
\vspace{-2mm}

We use this dataset to illustrate the interpretability of our ALE solution. We present the full set of results over this dataset in the appendix. As a summary, we find that ALE based feedback improves accuracy with statistical significance compared to the raw training data (P-value is $0.02$ and $0.04$ for {\withinale} and {\crossale} respectively). Our baselines slightly (1-2\% on average) outperform ALE, although without statistical significance (i.e., we cannot accept the alternate hypothesis that these approaches are better). However, the interpretability of the ALE based solution makes it far more desirable for AutoML feedback.

\begin{figure}[t]
  \centering
  \begin{subfigure}[t]{0.48\linewidth}
  \centering
  \includegraphics[width=1.00\textwidth]{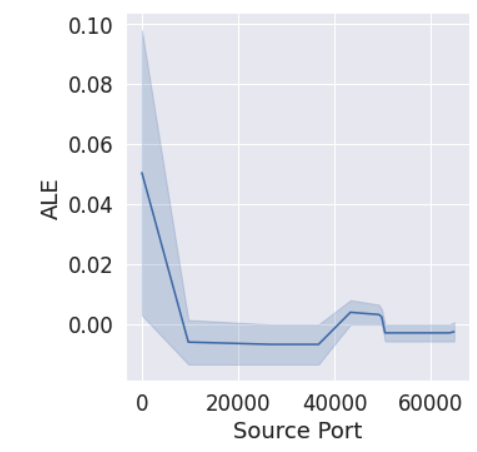}
  \caption{Unpredictive feature}
  \label{fig:ale_ex_1}
  \end{subfigure}
  \begin{subfigure}[t]{0.48\linewidth}
  \centering
  \includegraphics[width=1.00\textwidth]{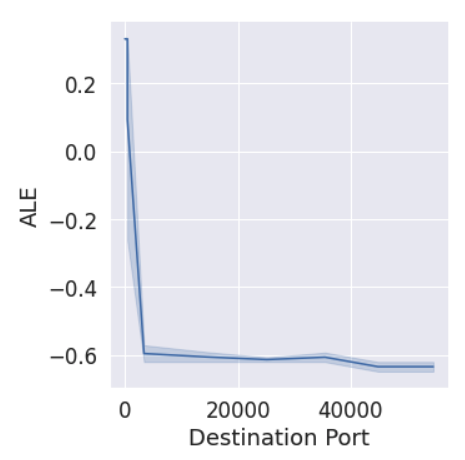}
  \caption{Predictive feature}
  \label{fig:ale_ex_2}
  \end{subfigure}
  \caption{Example ALE plots from the UCL dataset 2}
  \vspace{-3mm}
\end{figure}

We illustrate the true benefit of the interpretability of ALE in Figures~\ref{fig:ale_ex_1} and \ref{fig:ale_ex_2}. As Figure~\ref{fig:ale_ex_1} shows, the source port feature shows high variance especially around lower values. Based on domain knowledge, the AutoML user knows that networking source ports are typically assigned by the host kernel and therefore, though sometimes informative, are expected to be more noisy than other features. On the other hand, we see high variance across the destination port range 443-445 in Figure~\ref{fig:ale_ex_2}. Port 443 is reserved for HTTPs which is one of the most common targets of distributed denial-of-service (DDoS) attacks~\cite{privateeye}. Given these two insights, the user can discard the bound on the source port feature and focus on collecting more data around the 443-445 destination port range. In contrast, existing active learning solutions only choose a set of data points for labeling, and the AutoML user (1) has no idea why these points are needed; and (2) cannot leverage domain knowledge to determine the weaknesses of original data.   

\noindent \textbf{Setting the threshold.} Our solution, for the most part, is hyper-parameter free. The one exception is the threshold $\mathcal{T}$. In our experiments we used the median of the standard deviation across features: $0.02$ for the scream vs rest and $0.01$ for the two UCL datasets. Lower thresholds result in larger feature subspaces --- a larger area for the user to sample --- whereas larger thresholds result in smaller subspaces that are targeted. It is important to consider the users' sampling budget to set the threshold. When the budget is high, it is better to set the threshold to be lower: larger feature subspaces are more likely to help prevent overfitting. In contrast, when the sampling budget is low, a higher threshold may be better as it helps focus sampling on feature subspaces that are more likely to fall on the decision boundary.

\vspace{-3mm}
\section{Discussion and Limitations}
\vspace{-2mm}

Our work presents, to the best of our knowledge, the first interpretable feedback solution for AutoML. One of the benefits of the ALE-based feedback is that {\em users can prioritize bounds containing features they know can influence the label}. Users can also tune the threshold they use for each feature to be different based on their domain knowledge. Our approach also comes with the following limitations:

\noindent{\textbf{Bounded, at best, by the theoretical guarantees of QBC.}} Our feedback solution is based on QBC. While we do not provide a theoretical analysis of our solution, in principle, the theoretical analysis of QBC should apply here as well. However, it needs to be extended to account for our use of ALE instead of predictions as well as for the fact that we apply the feedback en-mass (as opposed to sample by sample). Part of the challenge in extending the theory to apply to this setting, is accounting for the fact that the ALE-feedback is not gradual and therefore the models in the AutoML ensemble need not fall in the version space after individual feedback samples are added to the training set. This limitation is similar to prior QBC work that employs simplified strategies to create the QBC committee such as boosting/bagging~\cite{DBLP:conf/icml/AbeM98} or sampling from a low dimension space~\cite{DBLP:conf/nips/Gilad-BachrachNT05}.  

\noindent{\textbf{Not applicable to AutoML systems that do not return an ensemble of models.}} Our solution takes advantage of AutoML systems such as AutoSKlearn~\cite{AutoSklearn} and TPoT~\cite{tpot} that return an ensemble of ML models. Some of the commercial AutoML offerings such as~\cite{azure-automl} also return an ensemble of models. Thus, there exist a sufficiently diverse set of AutoML systems where our solution is applicable. However, there are many AutoML systems where this is not the case e.g.,~\cite{DBLP:books/sp/19/SteinrueckenSJLG19}. In such cases, we would need an automated mechanism to find a sufficiently diverse set of ML models before we can return feedback to the user. This is a topic of future research. 

\vspace{-2mm}
\section{Related Work}
\vspace{-2mm}

\textbf{Active learning.} 
Decades of research in active learning established many strategies to suggest data for labeling (sometimes called "query learning") so that the learning algorithm can perform better with less training~\cite{activelearningsurvey}. Popular active learning strategies include uncertainty sampling~\cite{DBLP:conf/sigir/LewisG94}, query-by-committee~\cite{qbc}, expected model change~\cite{DBLP:conf/nips/SettlesCR07}, expected error reduction~\cite{DBLP:conf/icml/RoyM01}, and variance reduction~\cite{DBLP:conf/icdm/ZhengP02}. More recently, many works propose learning active learning strategies~\cite{DBLP:conf/icml/BaramEL03, DBLP:conf/aaai/HsuL15, DBLP:conf/icdm/ChuL16, DBLP:conf/nips/KonyushkovaSF17,DBLP:conf/emnlp/FangLC17,DBLP:conf/iccv/SinhaED19}. While we share similar goals, our algorithm is specifically designed for the users of AutoML without much ML expertise. Hence, our algorithm differs from prior active learning work in two major ways: (1) it provides interpretable data suggestions so that the users can understand the weaknesses of the training data; and (2) it naturally supports dynamic ML model and hyperparameter selection, which is the major focus of AutoML systems.

\textbf{AutoML.}
Driven by the explosive demand on ML solutions, AutoML aims to automate various aspects of ML developments so that domain experts can build ML applications without much requirement for ML knowledge~\cite{DBLP:journals/corr/abs-1904-12054}. AutoML has been a rapidly growing research area both in academia and industry~\cite{DBLP:journals/corr/abs-1908-00709}, and large companies already offer various AutoML solutions (e.g.,~\cite{google-automl, azure-automl, sagemaker-autopilot}). The goal of existing AutoML work is to automate the ML development pipelines for an input dataset, such as data cleaning/augmentation (e.g.,~\cite{DBLP:journals/pvldb/KrishnanWWFG16, DBLP:journals/corr/abs-1805-09501}), feature engineering (e.g.,~\cite{tpot, DBLP:books/sp/19/SteinrueckenSJLG19}), and model generation (e.g.,~\cite{AutoSklearn, DBLP:conf/icml/PhamGZLD18}). When the input data is insufficient, there is no established feedback. Our work takes a first step in this direction by focusing one specific type of feedback (new data suggestion), and complements AutoML systems.

\textbf{Interpretable machine learning.}
Interpretable ML aims to enable humans to understand the cause of a decision by ML models~\cite{DBLP:journals/corr/Miller17a}. They broadly fall into two categories: transparent models (e.g., decision trees) and post-hoc explanations (e.g., model-agnostic methods)~\cite{DBLP:journals/cacm/Lipton18}. 
In contrast, our work aims to provide interpretable new data suggestions to AutoML users, and we leverage a ML interpretation method (ALE) to provide such suggestions. 
\vspace{-2mm}
\section{Conclusion}
\vspace{-2mm}

The lack of feedback in AutoML systems is a major obstacle for non-ML experts to achieve acceptable performance. Our work takes a first step to address this challenge with an automated and interpretable feedback solution for AutoML. Our solution allows non-ML experts to identify data they can add to the training set to improve the accuracy of AutoML. Our key idea is to re-purpose a model-agnostic model interpretation solution, Accumulated Local Effect (ALE), to provide interpretable feedback. We use the variance across the ALE values for the models in an AutoML ensemble to identify areas where more data is needed. We show that our solution outperforms popular active learning solutions by up to $10\%$ when the user can freely collect additional samples based on the feedback, and the user can interpret our feedback on a per-feature basis. 



{
\balance
\bibliography{main}
\bibliographystyle{icml2021}
}

\clearpage

\onecolumn  

\appendix
\section{Results on UCL data set 2}
Tables~\ref{table::balanced_accuracy_UCL_2_within} and~\ref{table::balanced_accuracy_UCL_2_cross} show the results for within and cross ALE respectively. Although the average balanced accuracy across all algorithms seems similar at first glance, the Wilcoxon signed ranked test warrants pause: it indicates the ALE approach outperforms the no-feedback case and that none of the other baselines outperform it with statistical significance. The cross ALE approach in $64\%$ of the cases improves the balanced accuracy (the majority of the improvement ($48\%)$ is in the range $(0\%,10\%)$ where the $(a,b)$ is the set that does not include $a$ or $b$). Similarly, in $65.6\%$ of experiments the within ALE approach improves balanced accuracy (again in $36\%$ of the experiments the improvement is in the range (0\%,10\%)). As a means of comparison, QBC (with the same number of feedback points as the cross ALE approach) outperforms the original training set in $61.6\%$ of the cases. In fact, the cross ALE approach {\em strictly} outperforms QBC in $45.6\%$ of the experiments. We expect that higher gains in accuracy would have been possible if the user could have freely collected additional data points based on the feedback instead of focusing on the available candidate set.


\begin{table*}
\begin{center}
\begin{tabular}{ l c c c c}

\textbf{Algorithm (X)} & \textbf{balanced accuracy} & \textbf{P(no feedback, X)} &  \textbf{P(X, Within ALE)} & \textbf{P( Within ALE,X)}\\
\hline 
\hline
Without feedback & $91.0\% \pm 11.1\%$ & NA & $\textbf{0.02}$ & 0.97\\
{\color{blue}\textit{Within ALE}} & {\color{blue}$90.7\% \pm 11.7\%$} & {\color{blue} \textbf{0.02}} & NA & 0.97\\
Confidnece based & $92.4\% \pm 10.7\%$ & $0.02$ & $0.03$ & $0.96$ \\
QBC & $92.0\% \pm 10.9\%$ & $0.0002$ & $0.95$ & $0.04$\\ 
\hline
\end{tabular}
\end{center}
 \caption{UCL dataset 2 accuracy numbers for comparing wthin ALE with other algorithms. We have seperated the cross ALE from within ALE in this case to make sure the number of feedback points each algorithm uses is the same. For brevity we use P(x,y) to indicate the p-values of the one sided Wilcoxon signed ranked test that checks the hypothesis that $x == y$ and has the alternate hypothesis $x \le y$. Here we also show the reverse signed ranked test to demonstrate that baselines do not outperform ALE with statistical significance.\label{table::balanced_accuracy_UCL_2_within}}
\end{table*}

\begin{table*}
\begin{center}
\begin{tabular}{ l c c c c}

\textbf{Algorithm (X)} & \textbf{balanced accuracy} & \textbf{P(no feedback, X)} &  \textbf{P(X, cross ALE)} & \textbf{P( cross ALE,X)} \\
\hline 
\hline
Without feedback & $91.0\% \pm 11.1\%$ & NA & $\textbf{0.04}$ & $0.95$\\
{\color{blue}\textit{Cross ALE}} & {\color{blue}$91.7\% \pm 10.9\%$} & {\color{blue} \textbf{0.04}} & NA & $0.95$\\
Confidnece based & $93.1\% \pm 10.2\%$  & $0.02$ & $0.19$ & $0.80$\\
QBC & $93.4\% \pm 10\%$ & $0.0006$ & $0.55$ & $0.19$\\
\hline
\end{tabular}
\end{center}
 \caption{UCL dataset 2 accuracy numbers for comparing Cross ALE with other algorithms. We have seperated the cross ALE from within ALE in this case to make sure the number of feedback points each algorithm uses is the same. For brevity we use P(x,y) to indicate the p-values of the one sided Wilcoxon signed ranked test that checks the hypothesis that $x == y$ and has the alternate hypothesis $x \le y$.  Here we also show the reverse signed ranked test to demonstrate that baselines do not outperform ALE with statistical significance.\label{table::balanced_accuracy_UCL_2_cross}}
\end{table*}

\end{document}